\documentclass[sigconf]{acmart}
\AtBeginDocument{%
  }

\acmConference[28 Jan. 2025]{}{}{}

\begin{document}

\title{LLMs Provide Unstable Answers to Legal Questions}

\author{Andrew Blair-Stanek}
\email{ablair-stanek@law.umaryland.edu}
\affiliation{
  \institution{University of Maryland School of Law}
  \city{Baltimore}
  \state{Maryland}
  \country{USA}
}

\author{Benjamin Van Durme}
\email{vandurme@jhu.edu}
\affiliation{%
  \institution{Johns Hopkins University}
  \city{Baltimore}
  \state{Maryland}
  \country{USA}
}


\begin{abstract}
An LLM is ‘‘stable’’ if it reaches the same conclusion when asked the identical question multiple times. We find leading LLMs like gpt-4o, claude-3.5, and gemini-1.5 are unstable when providing answers to hard legal questions, even when made as deterministic as possible by setting temperature to 0.  We curate and release a novel dataset of 500 legal questions distilled from real cases, involving two parties, with facts, competing legal arguments,  and the question of which party should prevail.  When provided the exact same question, we observe that LLMs sometimes say one party should win, while other times saying the other party should win.  This instability has implications for the increasing numbers of legal AI products, legal processes, and lawyers relying on these LLMs.  

\end{abstract}




\maketitle

\section{Introduction}

One legal scholar has observed, ‘‘Flipping a coin to decide a case is among the most serious forms of judicial misconduct.’’ \citep{randomization_in_judging}  For example, Judge James Shull faced a hard case on child custody, which he resolved with a coin toss.  The Virginia Supreme Court permanently removed him from the bench \citep{shull_case}.  

We find that leading LLMs are like Judge Shull in that there is an element of chance in how they decide difficult legal questions.  Even if you take every precaution to make the calls deterministic, including setting the temperature to 0 and passing the exact same text of a legal question, leading LLMs will sometimes return that one party wins and sometimes return that the other party wins.  This random behavior is instability.  

The number and range of legal AI products built on the leading LLMs has increased rapidly, promising to automate ever more legal tasks and legal processes  \citep{gmi}. Ways have been proposed for LLMs to help judges decide cases and even directly resolve disputes \citep{LAI2024181, westermann2023llmediator}. The American Bar Association points lawyers to the leading LLMs to help with legal work, including brief writing  \citep{aba_LLMs, AI_Briefs}.  Most lawyers, judges, and law clerks using technologies built on LLMs assume that they are deterministic, like most computer programs.  

We introduce a novel dataset of 500 difficult legal questions,\footnote{All 500 questions and our code are at \url{https://github.com/BlairStanek/legal_instability/}} each question consisting of five paragraphs. These questions are distilled from published U.S. court decisions where the panel of judges split on which party should prevail.  Identifying facts like names are changed, since LLMs almost certainly saw the original cases during training.  There are only two possible answers to each question: either party~1 or party~2. 

We use our dataset to test the stability of three leading LLMs: gpt-4o, claude-3.5, and gemini-1.5.  For each of our 500 questions, we call each of the LLMs 20 times with the identical text and temperature=0.  We call an LLM unstable on a question if some of the 20 responses say party~1 should prevail, while other responses say party~2 should prevail.  We find all three LLMs are unstable on some percentage of our questions, ranging from 10.6\% for claude-3.5 up to 50.4\% for gemini-1.5.  Much of this instability is idiosyncratic to each model.   

We also test the stability of OpenAI's long-thinking LLM o1, even though o1 has a temperature hardwired at 1.0, making it inherently not deterministic.  We find o1 more unstable than claude-3.5 but less unstable than gpt-4o.  

We also measure each model's accuracy, with the groundtruth defined as the party that prevailed in the actual court case from which we distilled the question.  With statistical significance, we find gpt-4o and claude-3.5 perform slightly better than chance, while gemini-1.5 performs worse than chance.  

\section{Related Work}

Regarding terminology, both \textit{instability} and \textit{uncertainty} have been used in prior work to describe the  phenomenon of an LLM returning different answers when repeatedly passed identical text.  Throughout this paper, we use the term instability.  

There has been little research on LLMs' instability and none, to our knowledge, involving legal questions.  \citet{atil2024llm} use subsets of general-interest LLM benchmarks (e.g. European history) to measure stability.  They call LLMs with identical text, temperature=0, and a fixed seed for either 5 or 20 times to measure stability.  They observe instability leading to accuracy variations up to 10\%. \citet{lin2023generating} use several question-answering (QA) datasets to measure instability, albeit only with temperatures greater than zero. 
\citet{blackwell2024towards} use a benchmark set about cardinal direction reasoning to measure stability, with temperature=0. Their main focus is determining and quantifying the impact of instability on benchmark scores.  
\citet{liu2024your} introduce a new metric that blends both accuracy and stability, applying this metric on a new mathematical-reasoning dataset. \citet{renze2024effect} use tests from a wide variety of domains to confirm the expected result that setting a lower temperature for an LLM results in lower textual variance, as one would expect.  

\citet{lin2023generating} make a key distinction between stability and confidence.  Confidence describes how certain an LLM is that a \textit{single} answer it gives is correct, whereas stability is how dispersed an LLMs' answers are across \textit{multiple} calls.  Confidence is a concept long studied in statistics and machine learning  \citep{corbiere2019addressing,jiang2018trust,stats_book}.  One approach to evaluating confidence on a model without access to its parameters is to ask it to ascertain its confidence \citep{kadavath2022languagemodelsmostlyknow}. By contrast, we focus entirely on stability (and its absence, instability).  

Why are some leading LLMs unstable, even with temperature=0 and setting the seed?  It is impossible to say for sure, since the models are proprietary.  One possibility is nondeterminism in the ordering of floating-point accumulation \citep{pham2020problems}. 
Another is that the leading LLMs are hosted on cloud services, where calls to the same API may be handled by different servers, with different hardware with slightly different floating point implementations \citep{blackwell2024towards}.  Yet another is that API calls may be parallelized between multiple servers or processors, each with loads that vary over time, causing different orders of execution and thus changing the outcome.   

\section{Dataset Construction}

\subsection{Source Documents}

Our 500 difficult legal questions are distilled from 500 U.S. federal Courts of Appeal decisions where one judge dissented \citep{CAP}.  There are 13 Courts of Appeal, which are one level below the U.S. Supreme Court.  They typically hear cases in panels of three judges.  Often all three agree on the outcome of a case.  But in difficult cases, they can split 2-1, with one majority opinion and one dissenting opinion.  We use only these cases, which typically present a difficult legal question, with two well-written opinions that disagree on the proper resolution.   

In choosing all 500, we discard all cases with under 10,000 characters, as 2-1 cases of that length tend to hinge on simple procedural matters.  We also discard all cases with over 50,000 characters, as those tend to involve complex issues not easily summarized into our five-paragraph format.  

We also discard all cases involving diversity jurisdiction, which are lawsuits based on state law that end up in federal court (rather than state court) because the opposing parties are citizens of different states. We do this because each of the 50 states has its own body of law, and they sometimes diverge from each other.  Thus, we use cases based on federal law, which is largely uniform across all states. An important role of the U.S. Supreme Court, according to its own \href{https://www.law.cornell.edu/rules/supct/rule_10}{Rule 10}, is to resolve instances where federal law has become not uniform.

\subsection{Summarization and Name-Changing}

We summarize the 500 cases using OpenAI's o1 model, which we find excellent at summarizing cases while following detailed instructions.  Our instructions in the prompt tell o1 to change all person and entity names, other than the names of states, countries, and government agencies.  We do this because LLMs have almost certainly seen the original opinion, as U.S. federal court decisions cannot be copyrighted and are available from several internet sources.  

We tell o1 to create output with three paragraphs summarizing the relevant facts; one paragraph with the strongest legal arguments for one side; and one paragraph with the strongest arguments for the other side.  Our instructions to o1 tell it to present the two sides' best arguments with equal strength and not to hint how the Court of Appeals' majority ruled.  We then add the query, 

\begin{quote}Based on the facts above and your knowledge of the law, think step by step to figure out which party should prevail: <party~1> or <party~2>,\end{quote}

\noindent with the names of the two parties substituted here. This thinking step-by-step follows \cite{kojima22large} and elicits longer chains of reasoning and better end results.  

\subsection{Dataset Characteristics}

Using the above, we generate 500 difficult legal questions with the same five-paragraph format.  An example of one of these 500 is reproduced in full in Appendix~\ref{appendix_sample}.  The mean length is 665.4 words and the median is 657.0 words.  The standard deviation is 89.2 words, with a minimum of 439 words and a maximum of 963 words.  The length distribution neatly follows a normal distribution.  


Our code could be used to generate substantially more than our 500, as there are approximately 13,000 2-1 Court of Appeals decisions since 1950 meeting the criteria discussed above, and all are freely available in the public domain.  

\section{Experimental Setup }

We test on the three leading LLM foundational models: OpenAI's gpt-4o (specifically gpt-4o-2024-11-20), Anthropic's Claude-3.5   \linebreak (claude-3-5-sonnet-20241022), and Google's Gemini-1.5 (gemini-1.5-pro-002).  For the main experiments, we do not use OpenAI's o1, which has its temperature hardwired at 1.0.  With temperature 1.0, one would expect differing answers in multiple calls to the model with the same prompt. 

All calls are by API.  For all three models we run against, we set temperature to 0.0, which should maximize how deterministic the model is.  We also set all other available possible parameters to maximize how deterministic the model is. For gpt-4o, we fixed the seed.  We set $top\_p$ to 1.0 for gpt-4o and claude-3.5, and $top\_k$ to 1.0 for gemini-1.5.  

For each of the three models, we call the model 20 times with each of the 500 questions.  So, we make $500 \times 20 = 10,000$ calls against each model.  All experiments are 0-shot.  

Because of the part of the questions calling to ‘‘think step by step’’, the initial output of the LLMs is often lengthy legal analysis with headers and subheaders.  We thus always make a second call with the text ‘‘To summarize, which of the two parties do you think should prevail’’ to extract a clear single answer.  

\section{Results}
\subsection{Instability}

\begin{table}[t]
    \centering
    \caption{Proportion of questions for which each model was observed to be unstable (i.e. had stability less than 100\%) and Percent Unstable with 95\% Confidence Range.}
    \label{tab:stability}
    \begin{tabular}{lcc}
        \toprule
\textbf{Model} & \textbf{Number Unstable} & \textbf{Percent Unstable} \\
\midrule
Claude-3.5 & \phantom{0}53 / 500  & 10.6 $\pm$ 2.7 \\
GPT-4o & 215 / 500  & 43.0 $\pm$ 4.3 \\
Gemini-1.5 & 252 / 500  & 50.4 $\pm$ 4.4 \\
        \bottomrule
    \end{tabular}
\end{table}

We measure a model's \textbf{stability}, when passed the same text repeatedly, by taking the number of times the most-returned answer comes back, divided by the total number of calls.  For example, if we pass the same question 20 times to an LLM, and it returns that party2 should win 14 times and that party1 should win 6 times, the stability is $14/20=70\%$.  Since our problems have a binary outcome (either party1 or party2), stability ranges between 50 and 100\%.  We say that, on a particular question, a  model is \textbf{unstable} if and only if we observe stability less than 100\% on that particular question.

\begin{figure*}
  \centering
  \includegraphics[width=\linewidth]{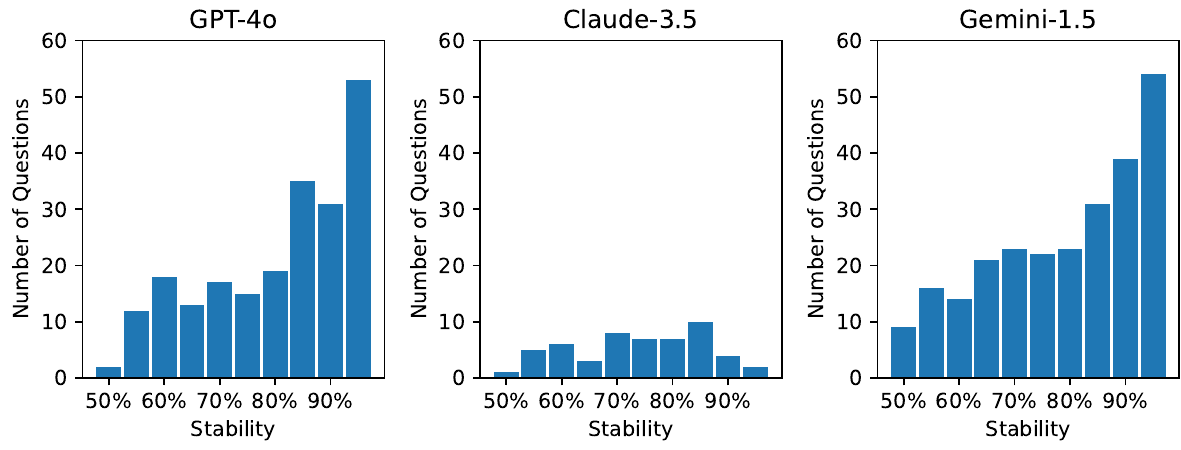}
  \caption{Histogram of the stability of all questions where stability was less than 100\% (i.e. the model was unstable).}
  \Description{Histogram for 3 models of count of problems with each possible observed instability, between 50 and 95 percent.}
  \label{fig_histogram}
\end{figure*}

The results of posing each of the 500 questions 20 times to each model are in Table~\ref{tab:stability}.  We see that claude-3.5 is observed to be unstable on the fewest number of questions, followed by gpt-4o, then gemini-1.5.  Histograms of the observed instability of those questions where the models were unstable are in Figure~\ref{fig_histogram}.  Both gpt-4o and gemini-1.5 have disproportionately more questions with observed stability just under 100\%, whereas claude-3.5 does not.  Beyond that, there are no clear patterns to the stability distribution.

\begin{figure}[H]
  \centering
  \includegraphics[width=2in]{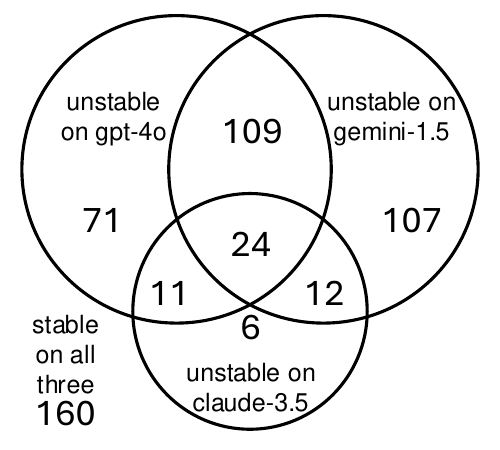}
  \caption{Venn diagram of which of the 500 questions each model or combination of models is unstable.}
  \label{fig_Venn}
  \Description{Three circles overlapping with numbers in intersections to reflect results.}
\end{figure}

\begin{table}[t]
    \centering
    \caption{Pairwise Pearson correlation of stability between the three models on the 500 questions.  The correlations are low, indicating that most instability is model-specific.}
    \label{tab:stability_correlation}
    \begin{tabular}{lcc}
        \toprule
\textbf{Model Pair} & \textbf{Correlation} & \textbf{p-value} \\
        \midrule
        GPT-4o vs. Claude-3.5 & 0.142 & 0.002  \\
        GPT-4o vs. Gemini-1.5 & 0.181 & 0.000  \\
        Claude-3.5 vs. Gemini-1.5 & 0.123 & 0.006  \\
        \bottomrule
    \end{tabular}
\end{table}

There is surprisingly little overlap between the questions that cause instability. Figure~\ref{fig_Venn} shows the distribution of the 500 questions among the set of those unstable on each of the three LLMs we test.  If the models' instability were completely independent, we would expect $500 \times 10.6\% \times 43.0\% \times 50.4\% = 11.5$ questions to be unstable with all three models. We see only twice that, with 24 questions causing instability for all three.  (One of these 24 is reproduced in full in Appendix~\ref{appendix_sample}.)  Gemini-1.5 has many questions on which only it is unstable; so does gpt-4o.  Looking at Table~\ref{tab:stability_correlation}, we see that the correlations between the stabilities of the three models is statistically significant (i.e. below 0.05) but small, ranging from 0.123 to 0.181.   Most of what makes the models unstable on a question appears to be idiosyncratic and specific to that model.  

The 24 questions that made all three models unstable spanned many areas of law.  Ten of the 24 involve criminal matters, mirroring the federal appeals courts' heavy criminal caseload.  Three focused on civil procedure, and there were two each for employment law and social security.  There was one case each for First Amendment, Native-American tribal jurisdiction, bankruptcy, civil rights, pensions, military, and immigration law. 

We considered whether stability decreases as length of the question increases.  The results are in Table~\ref{tab:length_vs_stability}.  The correlation is definitely not statistically significant for gpt-4o and claude-3.5, with the p-value being well over 0.05.  Indeed, the correlation for claude-3.5 is positive, which would mean it is more stable on longer questions, which makes little sense.  But for gemini-1.5 the correlation is very statistically significant and negative, meaning the longer the legal question, the less stable the model is.  

\begin{table}[b]
    \centering
    \caption{Pearson correlation between the question's stability and the question's length, measured by number of words.  Only gemini-1.5's result is statistically significant (i.e. below 0.05), whereas the other models' are not.}
    \label{tab:length_vs_stability}
    \begin{tabular}{lcc}
        \toprule
 & \textbf{Correlation} & \textbf{p-value} \\
\midrule
GPT-4o vs. length & -0.034 & 0.452 \\
Claude-3.5 vs. length & 0.024 & 0.590 \\
Gemini-1.5 vs. length & -0.121 & 0.007 \\
        \bottomrule
    \end{tabular}
\end{table}

Qualitatively, how does this instability arise?  We had a legal domain expert review pairs of answers from the same model on the same question, where the model concluded party~1 should prevail in one but party~2 should prevail in the other.  In none of them was the legal analysis unsound.  They simply weighed arguments or interests differently in different runs.  

For example, the question reproduced in Appendix~\ref{appendix_sample} made all three models unstable.  It involves the application of free speech principles to commercial activity; a key question is thus whether ‘‘the restriction be no more extensive than is necessary to serve the state interest.’’\footnote{Central Hudson Gas \& Electric Corp. v. Public Service Commission of NY, 447 U.S. 557, 572 (U.S. Supreme Court 1980).}  This standard is ambiguous. When the three models upheld the restriction, they found this standard met; when they found for the other side, they found this standard not met.  

\subsection{LLM Accuracy and Inter-LLM Agreement}

\begin{table}[b]
    \centering
    \caption{Agreement rate of the three LLMs with the court's outcome and with each other. One could view the rate of agreement with the court's actual outcome as accuracy. Confidence intervals are 95 percent.}
    \label{tab:agreement}
    \begin{tabular}{lcc}
        \toprule
  & \textbf{Percent Agreement} \\
\midrule
GPT-4o \& Court & 53.88 $\pm$ 0.98 \\
Claude-3.5 \& Court & 52.89 $\pm$ 0.98 \\
Gemini-1.5 \& Court & 46.44 $\pm$ 0.98 \\
\midrule
GPT-4o \& Claude-3.5  & 79.25 $\pm$ 0.79 \\
GPT-4o \& Gemini-1.5  & 71.72 $\pm$ 0.88 \\
Gemini-1.5 \& Claude-3.5 & 70.01 $\pm$ 0.90 \\
        \bottomrule
    \end{tabular}
\end{table}

The 10,000 calls to each LLM also provided an opportunity to measure their accuracy, with the groundtruth being the party that prevailed in the actual court opinion, with the majority of judges holding for them.  We would not expect accuracy anywhere near 100\%, given that actual domain experts (i.e. judges) split over which party should prevail.  But we would expect a good LLM to perform slightly over 50\%, for two reasons.  First, the arguments for the party that won the actual case are likely a bit stronger.  Second, since the U.S. has a common-law system, the court's decision for one party moved the law towards finding for similarly-situated parties.  

The top half of Table~\ref{tab:agreement} shows the accuracy of the three models, measured as agreement with the actual court's decision. Gpt-4o and claude-3.5's perform slightly over 50\%, with statistical significance, while gemini-1.5  performs worse.   

The bottom half of Table~\ref{tab:agreement} shows the pairwise agreement of the three models with each other.  The models all have higher agreement with each other than with the actual court opinion.  One possible explanation is that the summarization into five paragraphs biases each question in one direction, whereas the court had the benefit of full briefing and the full record.  Another explanation is that similar legal training corpora and training methods make the models tend to answer questions in the same way.  

\section{Experiments with o1}

Recall that we did not include o1 as one of the models being tested for stability because its temperature is hardwired as 1.0, which pushes a model to some instability.  Due to the high interest in o1 by the community, we do quantify o1's instability on our dataset.  Since o1 is expensive and since our experiments involve passing each question 20 times, we limit ourselves to calling one-tenth of our 500 questions.  We draw 50 questions evenly throughout the distribution of observed stability on o1's simpler relative gpt-4o.  
We used the same setup for calling o1 as we had for the other models, except that we obviously could not set temperature to 0 for o1.  

We found that o1 is unstable on 14 of the 50 questions, meaning 28 percent with a 95-percent confidence interval of $\pm 12$ percent. The probability that gpt-4o (with 215 of 500 observed unstable) and o1 (with 14 of 50 observed unstable) actually have the same underlying percent unstable is just 4\%, meaning o1 is 96\% likely to actually be more stable than gpt-4o.  We have several hypotheses for why.  Perhaps the deeper reasoning mechanism of o1 leads to more consistent outcomes.  Or, o1 API calls might be broken into parallel processes in different ways that increase stability.  Given the closed nature of both o1 and gpt-4o, we cannot test these hypotheses. 

\section*{Conclusion}

We find that gpt-4o, claude-3.5, gemini-1.5, and o1 all exhibit instability on some portion of the 500 questions in the new dataset we introduce.  These results suggest caution before using these LLMs in real-world legal applications.  

\section*{Limitations}

Further experiments could be performed with an even wider set of state of the art models. Cost and time make us focus on three industrial models at the time of this writing.  The LLMs we tested are proprietary, with ongoing development happening behind closed doors, meaning that we cannot reliably speculate how their APIs and underlying models may change going forward.  So, like many other works that explore such models, we cannot guarantee reproducibility in the future.

\section*{Acknowledgments}

This work has been supported by the U.S. National Science Foundation under grant No. 2204926. Any opinions, findings, and conclusions or recommendations expressed in this article are those of the authors and do not necessarily reflect the views of the National Science Foundation.

\bibliographystyle{ACM-Reference-Format}
\bibliography{sample-base}

\appendix
\section{Sample Question}
\label{appendix_sample}
Below is an example of one of the 500 questions in our dataset.  This question has 619 words. It was chosen since all three models had low stability on it: gpt-4o had observed stability of 0.60 on it, claude-3.5 had 0.55, and gemini-1.5 had 0.50.  Party~1 is a state government, while party~2 consists of the three individual plaintiffs who challenged a new policy of the state.  This question is available from GitHub in easy-to-copy text format \href{https://raw.githubusercontent.com/BlairStanek/legal_instability/refs/heads/main/DATASET/21_F.3d_1508.in.txt}{\underline{here}}.  This question was distilled from the actual case \textit{Lanphere \& Urbaniak v. State of Colorado}, 21 F.3d 1508 (10th Cir. 1994):  

\vspace{2ex}

You will be doing legal analysis of how you think a court should decide based on the facts below:

Several individuals in Redwood Falls, Colorado, including Mat\-thew Green and Ethan Brown, who formed the law partnership Green~\& Brown, and Richard Milton, who directed the Harbor Steps Recovery Program, had previously relied on certain criminal justice records to identify individuals facing charges for traffic and alcohol-related offenses. They used this information to send out direct mail advertising for professional services. The dispute arose when new legal restrictions were placed on accessing and using these particular records under a recently enacted state statute.

 According to court filings, the restriction barred release of names, addresses, and related information to persons intending to use those records for direct solicitations. The individuals from Green~\& Brown and Harbor Steps Recovery Program declined to sign the mandated statement prohibiting direct solicitation, which resulted in their being denied access to the records. They then challenged the statute, alleging that it violated their constitutional rights under the First and Fourteenth Amendments by preventing them from lawfully advertising to potential clients.

 Central to the underlying conflict was the question of whether there is a recognized right of public access to these kinds of criminal proceeding materials. Some positions in the dispute relied on traditions of open criminal processes in the United States, pointing out that many jurisdictions provide wide access to records used in criminal cases. Others emphasized that the records in question detailed accusations, not convictions, and that the public’s or a professional’s interest in contacting those listed must be balanced against individual privacy and the perceived risk of unwelcome solicitations. The record also reflected conflicting claims about whether the same information was obtainable from local news sources and about the significance of attorneys and treatment centers having prompt access for business-related purposes.

 One set of arguments insisted that no constitutional principle compels a state to disclose government-held data for direct commercial purposes. Supporters of this view referred to authorities such as Houchins v. KQED, Inc., which underscored that the Constitution does not guarantee wholesale access to all government information. They emphasized a legitimate state need to maintain privacy for individuals who have merely been charged, along with a desire to reduce potentially intrusive solicitations. They also contended that controlling record distribution helps safeguard public confidence in the criminal justice process, positing that unchecked commercial use of such information could lead to exploitation of vulnerable individuals. These arguments presented the statute as a tailored measure serving important governmental ends, pointing out that states traditionally enjoy discretion over whether or how to release official records.

 The other set of arguments maintained that open access to records on criminal charges has deep historical roots in the United States, as reflected by cases like Press-Enterprise Co. v. Superior Court and Globe Newspapers Co. v. Superior Court, which recognized a constitutional dimension to openness in criminal proceedings. From that vantage point, restrictions on these records were viewed as impermissibly burdening speech, particularly commercial speech that is still protected under precedents such as Shapero v. Kentucky Bar Ass’n and Zauderer v. Office of Disciplinary Counsel. Supporters of access argued that professionals like attorneys and treatment providers enhance public understanding by informing individuals of their rights and treatment options at critical junctures, thereby improving the criminal process overall. They also noted the possibility that those charged might not receive adequate counsel or needed services if such channels of communication were curtailed, undermining broader public interests in the fair administration of justice.

Based on the facts above and your knowledge of the law, think step by step to figure out which party should prevail: The State of Colorado or Green, Brown, and Milton

\end{document}